\documentclass[lettersize,journal]{IEEEtran}
\usepackage{amsmath,amsfonts}
\usepackage{graphicx}
\usepackage{subcaption}

\begin{document}

\title{Domain Generalization via Ensemble Stacking for Face Presentation Attack Detection}

\author{Usman Muhammad, Jorma Laaksonen, Djamila Romaissa Beddiar, and Mourad Oussalah, ~\IEEEmembership{Senior Member,~IEEE}
\thanks{Usman Muhammad and Jorma Laaksonen are associated with the Department of Computer Science at Aalto University, Finland. Djamila Romaissa Beddiar and Mourad Oussalah are affiliated with the Center for Machine Vision and Signal Analysis (CMVS) at the University of Oulu in Finland (e-mail: usman.muhammad@aalto.fi).}
\thanks{}}

\markboth{The work has been submitted to a peer-reviewed journal}%
{Shell \MakeLowercase{\textit{et al.}}: A Sample Article Using IEEEtran.cls for IEEE Journals}


\maketitle

\begin{abstract}
Face Presentation Attack Detection (PAD) plays a pivotal role in securing face recognition systems against spoofing attacks. Although great progress has been made in designing face PAD methods, developing a model that can generalize well to unseen test domains remains a significant challenge. Moreover, due to different types of spoofing attacks, creating a dataset with a sufficient number of samples for training deep neural networks is a laborious task. This work proposes a comprehensive solution that combines synthetic data generation and deep ensemble learning to enhance the generalization capabilities of face PAD. Specifically, synthetic data is generated by blending a static image with spatiotemporal encoded images using alpha composition and video distillation. This way, we simulate motion blur with varying alpha values, thereby generating diverse subsets of synthetic data that contribute to a more enriched training set. Furthermore, multiple base models are trained on each subset of synthetic data using stacked ensemble learning. This allows the models to learn complementary features and representations from different synthetic subsets. The meta-features generated by the base models are used as input to a new model called the meta-model. The latter combines the predictions from the base models, leveraging their complementary information to better handle unseen target domains and enhance the overall performance. Experimental results on four datasets demonstrate low half total error rates (HTERs) on three benchmark datasets: CASIA-MFSD (8.92\%), MSU-MFSD (4.81\%), and OULU-NPU (6.70\%). The approach shows potential for advancing presentation attack detection by utilizing large-scale synthetic data and the meta-model.
\end{abstract}

 \begin{figure}
	\centering
	\includegraphics[scale=.30]{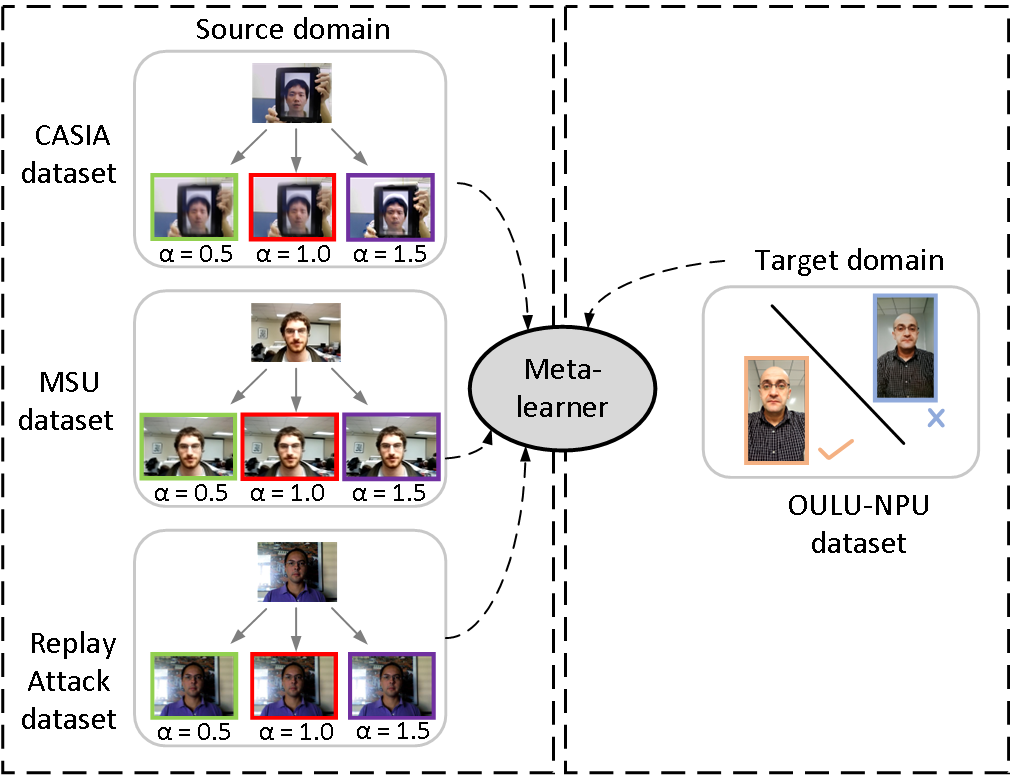}
	\vspace{0.4cm}
	\caption{Domain generalization: The source domains are trained with diverse sets of synthetic images, and the meta-learner leverages this diversity to acquire complementary information, enabling robust generalization to unseen target distributions.}
	\label{FIG:1}
\end{figure}
\section{Introduction}
\IEEEPARstart{O}{ver} the past few decades, facial recognition (FR) technology has been frequently used in numerous real-world applications, such as mobile payments, access control, immigration, education, surveillance, and healthcare \cite{kim2022structural}. The accuracy of FR is no longer a major concern, and the error rate has dropped to 0.08\%, according to National Institute of Standards and Technology (NIST) \cite{grother2018ongoing}. Despite its great success, a simple FR system might be vulnerable to spoofing, known as a presentation attack. For instance, print attacks, video replays, and 3D masks are the most common attacks reported recently in the face anti-spoofing domain \cite{muhammad2022face, muhammad2022selfb, wu2021dual, jia20203d, arashloo2020unseen}. Thus, several hand-crafted and deep representation methods have been proposed to protect FR systems against presentation attacks \cite{boulkenafet2016face, liu2021face, shao2020regularized, muhammad2022self, wang2020cross, saha2020domain, muhammad2019faceb, shao2019multi}. Many of these models reported promising performance in the intra-domain testing scenario. However, their performances remain limited in the cross-dataset testing scenario due to the distributional discrepancy between the source domain and the target domain.\\
\indent In the context of cross-dataset testing scenarios, a key contributing factor to the performance limitations of deep learning models can often be attributed to the restricted size or inadequacy of the training dataset. Another reason is rooted back to the inherent assumption that in many face presentation attack detection methods, training and testing data come from the same target distribution. This opens us several questions. For instance, if a model is trained on cut photo attack images, would it work on mask attack images? What if a model is trained only on replay attack images, but tested on warped photo attacks? Is it possible to deploy a model that is trained using different illumination conditions and background scenes under control lighting scenario? Answers to all these questions depend on how a machine learning model can deal with this domain shift problem. Therefore, domain generalization (DG) is essential when the objective is to build a model that can perform well on entirely new and diverse domains that were not seen during its training. As shown in Fig. 1, domain generalization refers to the task of training a model on data from multiple source domains and then deployed on a new, unseen target domain. To address the domain generalization issue, the majority of face anti-spoofing methods, such as adversarial learning \cite{liu2022adversarial}, meta pattern learning \cite{cai2022learning}, generative domain adaptation \cite{zhou2022generative}, hypothesis verification \cite{liu2022feature}, and cross-adversarial learning \cite{huang2022generalized}, have been introduced to improve the generalization ability of the model.\\
\indent Since generalization is a fundamental challenge in machine learning, researchers have explored various generalization-related research topics such as meta-learning (learning to learn), regularization techniques, ensemble learning, and data augmentation \cite{wang2020cross, saha2020domain, shao2019multi}. In particular, domain generalization is important in face anti-spoofing because collecting and annotating new datasets in real-world scenarios can be expensive and time-consuming. On the other hand, domain adaptation (DA) approaches focus on adapting a model trained on one or multiple source domains to perform well on a specific target domain that is different from the source domains. The main difference between these two approaches is that domain adaptation assumes that the target domain is known during training the model, while domain generalization does not. For instance, research in \cite{jia2020single} relies on a shared feature space and assumes that it would also be invariant to domain shift. However, this assumption has limitations in face PAD because when the source domains become more diverse, learning a domain-invariant model becomes more difficult \cite{zhou2021domain}. Instead of concentrating on some domain-specific cues such as the paper texture, diverse data can help the model generalize better if more generalized cues are shared by all source domains \cite{shao2019multi}. Additionally, spoofing attacks are launched physically by malicious hackers, outside the control of the biometric system. Therefore, domain generalization is more important than domain adaptation because DG does not require the target domain to be known during training.\\
\indent Recently, video-based methods \cite{shao2020regularized, wang2020cross, muhammad2023self} have shown potential to offer improved generalization compared to image-based methods \cite{boulkenafet2016face, wen2015face, muhammad2019face}. Nonetheless, not all frames in a video contain essential information, and processing every frame increases the overall computational load \cite{muhammad2022adaptive}. Thus, frame selection approaches are often used in PAD literature to identify frames with critical facial actions or dynamic cues, such as blinks, eye movements, or facial expressions, which are more indicative of a live person. For instance, the authors in \cite{muhammad2023face} claimed that selecting specific frames based on a Gaussian weighting function can capture different facial actions and generalize better to unseen spoofing attempts. Although adaptive frame selection strategies allow the model to focus on frames that contribute the most to the final result \cite{muhammad2023face}, maintaining temporal coherence in videos while using adaptive frame selection strategies can be a challenging task. Thus, more advanced techniques are needed to account for the motion of objects in the scene before selecting the optimal frames. The rank pooling is used to aggregate spatio-temporal features from video data into a fixed-length representation  \cite{yu2020fas}. However, rank pooling can be sensitive to noisy or outlier frames in the video, as it considers the rank order of features.\\
\indent To mitigate the impact of camera motion-induced noise, feature-based global motion estimation methods are introduced to compensate the camera motion \cite{muhammad2022self, muhammad2022adaptive}. During global motion estimation, the objective is to estimate a transformation (e.g., affine or homography) that aligns the source image with the target image. However, this transformation may cause parts of the target frame to remain empty, leading to black borders around the transformed image \cite{muhammad2022self}. This occurs because each successive transformation, such as rotations, translations, scaling, or affine changes, modifies the spatial characteristics of the image. These cumulative alterations can result in the transformed image exceeding the dimensions of the original bounding box. Although this issue is addressed by selecting dense sampling approach in \cite{muhammad2022adaptive}, the dense feature extraction significantly increases the computational cost of the model. Moreover, we argue that black framing can introduce artificial cues into the video data, which might mislead or bias the models during training. Thus, addressing black framing is essential before the training any models.\\
\indent Motivated by the aforementioned observations, we extend our previous approach \cite{muhammad2022self} in the following ways: (i) we introduce a new data augmentation technique as a post-processing step to composite the transformed image back into the target frame seamlessly, reducing the visibility of black framing artifacts, and helping the model learn to recognize actions in the presence of motion blur; (ii) we address the domain generalization issue by learning from the diversity of the proposed synthetic data and introducing a deep ensemble learning framework; (iii) we use several explainability methods to answer questions such as “why did the model make a particular prediction?” or “what features were most influential in the decision-making process?”; and (iv) we balance the computational cost of the global motion estimation and system performance.\\
\indent Intuitively, our proposed method has at least three main advantages. First, the video distillation technique allows the model to operate on a smaller subset of frames, reducing the computational overhead while maintaining or even improving generalization. Second, the static image, being a single frame, may not always provide the entire information, while dynamic cues, such as eye blinking, facial expressions, or head movement (i.e., tilting, turning, or nodding), are crucial to differentiate between real and attack samples. By blending static image with the spatiotemporal encoded image, we combine the stability of the spatiotemporal image with the smoother, less jittery motion of the object. This results in a visually improved representation that decreases the perception of motion blur. Third, we can easily control the sampling rate by adjusting the segment size of the video. This is important because a higher sampling rate results in a higher temporal resolution but potentially more noisy, while a lower sampling rate means less frequent sampling, leading to a smoother representation but potentially lower temporal detail.
In summary, our key contributions can be summarized as follows: 
\begin{itemize} 
\item We introduce a video-based data augmentation mechanism by considering both the spatial and temporal domains of the video. The proposed approach can assist deep learning models in capturing spatiotemporal information and enhancing their performance in face (PAD) tasks.
\item A meta-model is presented that leverages information from different subsets of synthetic samples, leading to improvements in the overall performance and robustness of the model.
\item Explainability techniques, which include gradient-weighted class activation mapping, occusion sensitivity map and LIME are employed to explain the decisions made by the employed model. The model reveals that motion cues are the most important factors for distinguishing whether an input image is spoofed or not.
\item Experiments on four benchmark datasets, consisting of CASIA-MFSD, Replay-Attack, MSU-MFSD, and OULU-NPU databases, show that our proposed method is significantly superior on three databases in comparison with other state-of-the-art generalization methods used now.
\end{itemize}

The rest of this work is organized as follows: Section II discusses recent developments and related past works. Section III explains all the steps of the proposed method. Section IV shows the implementation details, ablation study, and comparison against several public benchmark datasets. Section 5 concludes the entire work and gives suggestions for future research.
 \begin{figure*}
	\centering
	\includegraphics[scale=.41]{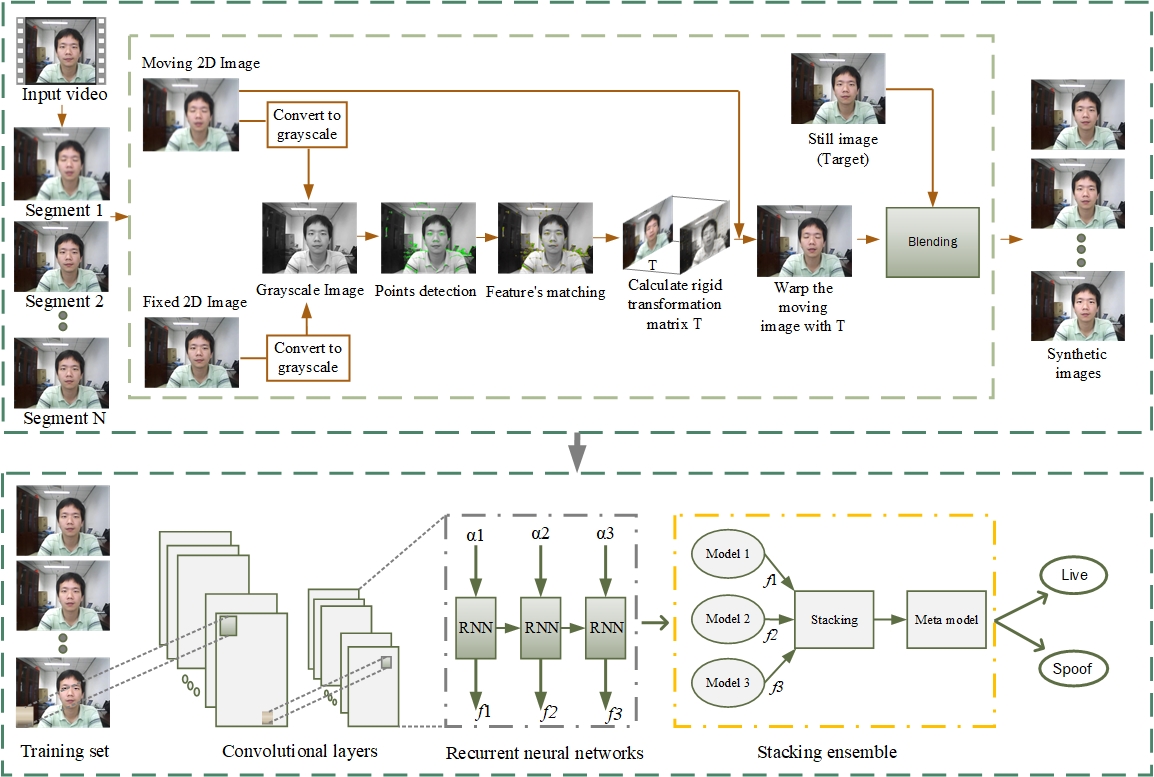}
	\vspace{0.4cm}
	\caption{Schematic diagram of the proposed data augmentation and deep ensemble learning for face anti-spoofing countermeasure.}
	\label{FIG:3}
\end{figure*}

\section{Literature Review}
Over the past few years, face PAD methods have received considerable attention from both academia and industry. In general, these methods can be roughly classified into appearance-based methods and temporal-based methods.\\
\indent {\em Appearance-based methods:} Traditional appearance-based methods usually extract hand-crafted features such as LBP \cite{freitas2012lbp} and SIFT \cite{patel2016secure} based on various appearance cues. The authors in \cite{boulkenafet2016face} claimed that color information is crucial and luminance-chrominance color spaces improve the detection performance of face PAD in comparison to the RGB and the gray-scale image representations. The multiscale filtering approach proposed in \cite{boulkenafet2016scale} was found to be effective and LBP-based multiscale features provided improved performance. Wen {\textit{et al}} \cite{wen2015face} utilized image distortion analysis (IDA) and developed an ensemble classifier, where multiple SVM classifiers were implemented. In particular, the features were selected based on specular reflection, blurriness, chromatic moment, and color diversity to provide input to the SVM classifiers. A component-based coding framework was proposed to encode different components of the face in \cite{yang2013face}. To deploy secure face locking on a smartphone, a method was developed based on extracting color distortion, Moiré-pattern analysis, surface reflection, and shape deformation \cite{patel2016secure}. The LBP features were combined with the feature maps of a deep learning model to improve the detection of face PAD in \cite{li2019face}. The authors showed that the need for large training samples in face PAD can be mitigated by using convolutional feature maps. A hybrid deep learning method was introduced in \cite{muhammad2019face} to encode appearance information from two CNNs and the SVM classifier was used to discriminate live and spoofed images. Although appearance-based methods provide improved performance in an intra-database testing scenario, the performance remains limited when evaluated on a completely unseen testing domain.\\
\indent {\em Temporal-based methods:} The study reported in \cite{muhammad2022self} estimates global motion and amplifies motion cues such as hand movements or head rotation where BiLSTM \cite{schuster1997bidirectional} is used to predict the motion. However, the study provides black framing at the edges of encoded images which was successfully addressed by employing dense sampling along with similarity transformations, as reported in \cite{muhammad2022adaptive}. Moreover, in order to encode head movements, eye-blinking, and lip movements, a dynamic mode decomposition (DMD) method has been introduced to capture the temporal cues from frame sequences \cite{tirunagari2015detection}. Eulerian motion magnification was used to magnify the facial expressions in \cite{bharadwaj2013computationally}. Then, local descriptors, such as HOOF, and LBP were utilized to improve the classification performance \cite{bharadwaj2013computationally}. The Photoplethysmography (rPPG) signal was found to be crucial to improve the face PAD performance \cite{liu2021multi}. A unified framework based on CNN-BiLSTM was used to capture both appearance and temporal cues in \cite{muhammad2019face}. A study conducted in \cite{sun2007blinking} showed that the spontaneous blinking of a person provides an intrinsic detection cue to improve live face detection. A dense optical flow scheme is proposed to estimate the motion of two successive frames in \cite{yin2016face}. The authors claimed that real and attack videos have different optical flow motion patterns, which help to improve the PAD performance. A 3D CNN model ass employed to capture both spatial and temporal information in \cite{li2018learning}. A combined CNN-RNN model was developed to capture the auxiliary information (i.e., the depth map and rPPG signals) for improving the detection performance \cite{liu2018learning}. However, when the temporal and appearance-based methods were employed in a cross-dataset scenario, the detection performance remained vulnerable to degradation due to real-world variations (such as user demographics, input cameras, and variations in illumination). Therefore, domain generalization that aims to learn from several source domains becomes significant while dealing with presentation attack detection. \\
\indent {\em Deep Domain Generalization methods:} Several deep domain generalization methods have been introduced to improve the generalization ability of face PAD. For instance, a domain adaptation method that generates pseudo-labeled samples named cyclically disentangled feature translation network (CDFTN) was proposed in \cite{yue2022cyclically}. Chuang {\textit{et al}} proposed to improve the generalization based on one-side triplet loss \cite{chuang2022generalized}. In \cite{cai2022learning}, a two-stream network was utilized to fuse the input RGB image, while a meta-pattern learning was proposed to improve the generalization. In \cite{huang2022generalized}, a cross-adversarial training scheme was proposed to improve the generalization by minimizing the correlation among two sets of features. The work reported in \cite{zhou2022generative} aims to learn a generalized feature space by designing the target data to the source-domain style called generative domain adaptation (GDA). A hypothesis verification framework was proposed in \cite{liu2022feature} where two hypothesis verification modules are utilized for improving the generalization. In \cite{wang2022domain}, a novel shuffled style assembly network (SSAN) was introduced by aligning multiple source domains into a stylized feature space, while domain generalization was improved by a contrastive learning strategy. To select a common features space, adversarial learning was proposed and aggregation of live faces was performed to achieve a generalized feature space in  \cite{liu2022adversarial}. Nonetheless, there is no clear consensus that the pre-defined distributions can be considered the optimal ones for the feature space. We argue that different domains have their strengths and weaknesses. By combining them, their collective knowledge and diversity can lead to a more comprehensive understanding of faces. However, implementing and fine-tuning such an approach might be complex and requires careful consideration of alignment methods, ensemble learning techniques, and evaluation strategies to ensure the desired performance improvements are achieved.

\section{Methodology}
A key idea underlying our approach is to enhance the model's generalization ability. This objective is achieved through training multiple sub-models using distinct subsets of synthetic data. Specifically, we embark on data augmentation by blending a static image with spatiotemporal encoded images, effectively replicating the visual effect of motion blur through the manipulation of alpha values applied to static image. This variation in alpha values governs the extent of each image's influence on the final composite, thereby yielding an array of synthetic data subsets. As a further stride in our methodology, we introduce a meta-model, which capitalizes on the predictions of the base models. In the following sections, we describe all these steps in detail. An overall view of our proposed methodology is presented in Fig. 2.

\subsection{Data augmentation}
The process of data augmentation involves five main steps as follows: (1) Select a source and a target image; (2) Estimate the global motion transformation between the source and target images; (3) Perform the geometric transformation on the source image to align it with the target frame; (4) Create a new spatiotemporal encoded image by stabilizing the segments of the video; and (5) Use alpha blending to combine the transformed image with the target (static) frame, considering the transparency of each pixel. Each of these steps are explained in the following sub-sections.

\subsubsection{Global motion estimation}
Suppose a video $V$ is equally divided into $P$ non-overlapping segments, i.e., $V = \{S_k\}^P_{k=1}$, where $S_k$ is the $k$-th segment. The length of each segment is set to be $N=40$ frames. For each segment, features are extracted from the fixed (first) and moving (second) images of the segment. In particular, the FAST feature detector \cite{rosten2005fusing} is utilized to detect interest points, and then the FREAK descriptor \cite{alahi2012freak} extracts the features to collect points of interest from both frames. For matching the interest points, Hamming distance (HD) is utilized in our work.
The transformation between frames is calculated from the first frame onward using a rigid (Euclidean space) transformation. The rigid transformation preserve lengths, angles, and shapes, and include, translation, rotation, and reflection. On the contrary, the affine transformation used in \cite{muhammad2022self} include, translation, rotation, scaling, and shearing, and preserves parallel lines and ratios of distances, but can change shapes through scaling and shearing. Thus, rigid transformation for each subsequent frame in the segment is estimated as:
\begin{equation}
\begin{bmatrix} a' \\ b' \\ 1 \end{bmatrix} = \begin{bmatrix} d_{11} & d_{12} & d_{13} \\ d_{21} & d_{22} & d_{23} \\ d_{31} & d_{32} & d_{33} \end{bmatrix} = M\begin{bmatrix} a \\ b \\ 1 \end{bmatrix}
\end{equation}
In this equation, $\begin{bmatrix} a' \ b' \ 1 \end{bmatrix}$ represents the homogeneous coordinates in the fixed image, and $\begin{bmatrix} a \ b \ 1 \end{bmatrix}$ represents the homogeneous coordinates in the moving image. The rigid transformation matrix $M$, which describes the inter-frame motion. To eliminate false-matching points and robust estimation of the geometric transformation between the frames, we use the M-estimator Sample Consensus (MSAC) algorithm \cite{torr1997robust} to detect outliers and remove false matching points. To obtain a warped (spatiotemporal) image, accumulation is performed using the following equation \cite{muhammad2022adaptive}:
\begin{align}
A = \frac{1}{N} \sum_{i=1}^{N} I_i \cdot M,
\end{align}
where $A$ is a single image that represents the combined motion in the segment after applying the accumulated transformation $M$. $N$ is the number of frames that are considered for accumulation. $I_i$ represents the $i-$th frame in the segment. Each frame $I_i$ is transformed using the accumulated transformation $M$ before being added to the sum. $\frac{1}{N}$ is the scaling factor that ensures the final result is an average. In particular, Eq. 2 shows that we take each frame in the segment, apply the accumulated transformation $M$ to it, sum up these transformed frames, and then divide by $N$ to generate a final composite frame $A$ for that particular segment. To show the importance of this approach, we simply remove the accumulated transformation $M$ and use the following equation:
\begin{align}
B = \frac{1}{N} \sum_{i=1}^{N} I_i,
\end{align}
The equation calculates the average frame by summing all the individual frames in the segment and then dividing that sum by the total number of frames $N$. This results in the average frame of the video. The outcome of both equations is illustrated in Fig. 3. The first row represents the noisy camera motion, which can be observed after using the Eq. 3. The second row shows that the proposed approach significantly removes the motion blur using the Eq. 2. Although the camera motion issue is eliminated, black framing near the border of the images in the second row appears, which requires further preprocessing. In the following sub-section, we address this issue and explain the motivation of data augmentation. 
\begin{figure}
\centerline{\includegraphics[width=7.00cm]{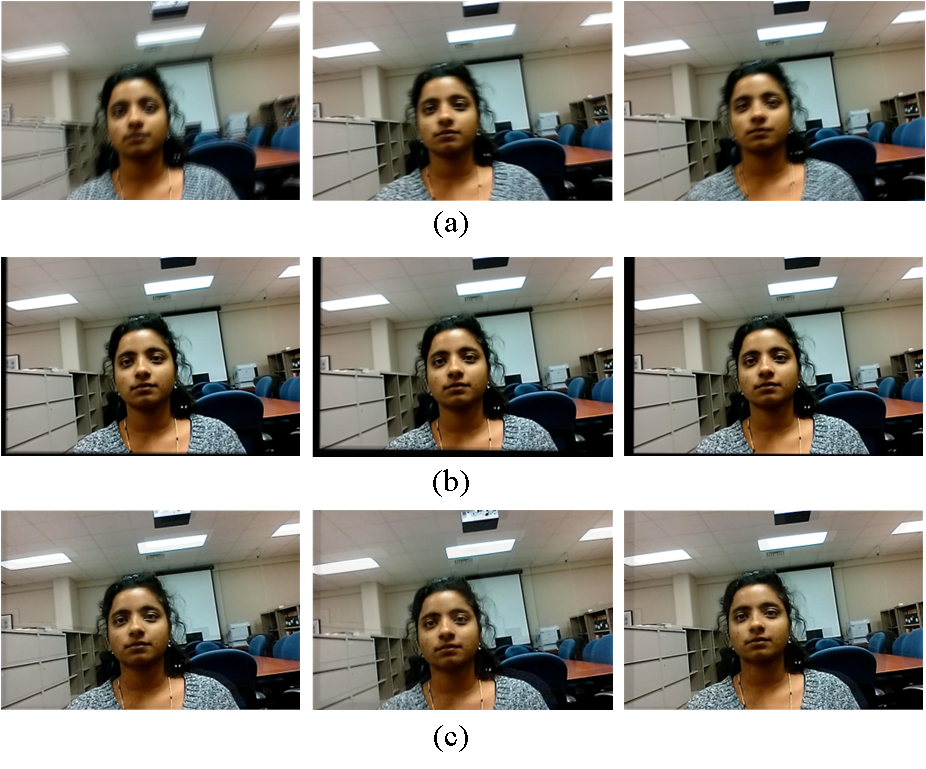}}
\caption{(a) Temporal averaging is performed to visualize the global motion, which exhibits a significant amount of distortion in the encoded image. (b) The cumulative transformation is calculated by applying the transformation matrix of each frame's motion to the previous cumulative transformation. (c) The result of the cumulative transformation with alpha blending.}
	\label{FIG:4}
\end{figure}

\subsubsection{Alpha transparency}
While Eq. 2 effectively removes the disruptive effects of noisy camera motion, the spatiotemporal-encoded images inherently exhibit a certain level of motion blur. This blur arises from the inherent movement between the camera and the scene during each exposure. In particular, if the subjects move their hands or change the direction of their face in the scene while the camera's shutter is open, their motion appears as a blur in the spatiotemporal-encoded image. This observation has encouraged us to propose a new basic augmentation technique, called \textit{alpha transparency}, for face PAD. The concept is to assign appropriate alpha values to the first (static) frame of the segment and then blend it with the spatiotemporal-encoded image based on those alpha values. This approach helps preserve more details and fine features of the scene, as the spatiotemporal-encoded image captures different stages of the motion blur. The alpha blending process is achieved through the following two steps: (1) Obtain a source image (i.e., a spatiotemporal-encoded image based on Eq. 2); and (2) Choose first (still) image of each segment to blend with the source image. Let us assume, without loss of generality, that we blend the source image $P_1$ over the target image $P_2$ as follows:
\begin{align}
P_{blended}(a,b) = \alpha P_{1}(a,b) + (1-\alpha)P_{2}(a,b).
\end{align}
The equation represents how to combine the information from the two source values $P_{1}(a,b)$ and $P_{2}(a,b)$ to create a new value at the location $(a,b)$ in the synthetic image. The weight factor $\alpha$ determines the contribution of $P_{1}(a,b)$ to the blended result. When $\alpha$ is closer to $1$, $P_{1}(a,b)$ has a higher influence, and when $\alpha$ is closer to $0$, $P_{2}(a,b)$ has a higher influence. Thus, blending the source image with the target image helps to eliminate the black framing issue as shown in third row of Fig. 3, creating a visually seamless transition. Moreover, Fig. 4 displays images that represent the results of blending with different $\alpha$ values. This approach can be useful in scenarios where someone wants to emphasize the overall motion of the scene while maintaining a recognizable background or main subject. We hypothesize that the combination of alpha blending and spatiotemporal encoding can be beneficial in simulating motion blur and help the deep learning model become more robust to real-world scenarios where camera motion is present.
  \begin{figure}
	\centering
	\includegraphics[scale=.30]{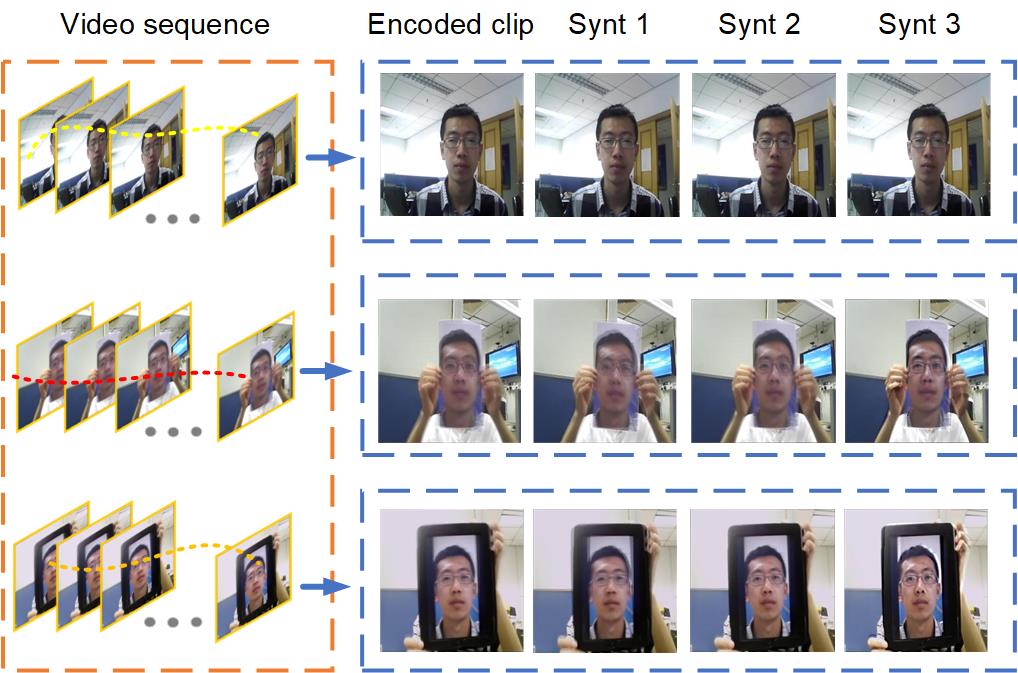}
	\vspace{0.4cm}
	\caption{2D synthetic samples from CASIA-MFSD are shown. In the left column, we present video segments used in the process of data augmentation. In the right column, we display still images after blending with spatiotemporal images using alpha values of 0.5 (Synt 1), 1.0 (Synt 2), and 1.5 (Synt 3), respectively. These synthetic samples can be effectively employed for ensemble stacking, resulting in a significant improvement in face anti-spoofing performance.}
	\label{FIG:2}
\end{figure}

\subsection{Deep ensemble learning}
Deep learning methods based on 2D Convolutional Neural Networks (CNNs) have shown improved performance compared to classical machine learning approaches \cite{wang2020cross,liu2021face,shao2020regularized}. However, mainstream 2D CNN frameworks primarily focus on spatial information, thus lacking the capacity to understand sequential data. Moreover, 2D CNNs do not possess a memory mechanism to capture temporal relations. Motivated by the fact that recurrent neural networks (RNNs) can effectively handle temporal information, we develop a stacking-based deep ensemble learning framework to learn from motion blur by processing sequences of images representing motion over time. Ensemble learning has been supported by multiple approaches such as bagging, boosting, and stacking, resulting in a better generalization of learning models \cite{fatemifar2020stacking}. Specifically, stacking is an integration technique that involves combining the predictions based on different weak model predictions, where the meta-model is used to integrate the output of base models \cite{ganaie2021ensemble}. One common approach in stacked ensemble learning is to develop a set of Tier-1 classifiers denoted as $S_{1}, S_{2}, S_{3}, \ldots, S_{N}$. These classifiers are developed through cross-validation of the training dataset. Then, the outputs or predictions from these Tier-1 classifiers are subsequently employed as inputs for a higher-level classifier or meta-learner within the ensemble \cite{polikar2012ensemble}.\\
\indent Since our primary goal is to learn from the variations of motion blur sequences, we train diverse recurrent neural networks to leverage the benefits of the proposed data augmentation mechanism. The approach involves fine-tuning a CNN model on the labeled datasets in the first stage. Then, we extract the fine-tuned features from the pooling layer and utilize them as input to train different variants of RNNs. In our work, three base models—Long Short-Term Memory (LSTM) \cite{hochreiter1997long}, Bidirectional Long Short-Term Memory (BiLSTM) \cite{schuster1997bidirectional}, and Gated Recurrent Unit (GRU) \cite{cho2014learning} are selected. The first base model (LSTM) captures temporal patterns and dynamics across frames using its special memory cells, allowing it to maintain long-range dependencies in sequential data. When presented with a sequence of blurred frames, the LSTM can learn to recognize patterns one by one and retain relationships between consecutive frames. The second base model (BiLSTM) captures information from both past and future contexts in sequential data. It consists of two sets of LSTM cells, where both cells work independently to capture information from both past and future contexts. The hidden states from both directions are then concatenated or combined to obtain the final output. The third base model (GRU) incorporates gating mechanisms similar to LSTM, enabling it to selectively learn and update information over time.\\
\indent To train RNNs for learning from motion blur sequences, we utilize $k$ cross-validation, where each synthetic sample in the dataset gets to be in the training set $k-n$ times. This approach ensures that the model's performance is evaluated on the entire dataset without introducing data leakage or bias due to a single train-test split. To better understand the learning of sub-models, Fig. 5 shows the proposed validation scheme, where each RNN is trained with $k-1$ folds, $k-2$ folds, and $k-3$ folds to fully utilize stacking. This means that we create multiple experts by training each model on different training subsets. Consequently, each model is reinforced to concentrate on different aspects of data, such as variations in temporal blur. For instance, one model can focus on certain types of features using a subset of synthetic data, while another model may perform better on other aspects. We then combine the predictions from these expert (base models) and use them as input to a meta-model (another RNN). The meta-model learns to combine these predictions and make the final prediction. We call it a meta-model because it leverages the diversity and complementary strengths of individual base models, leading to improved generalization for face PAD.\\
\begin{figure}
\centerline{\includegraphics[scale=.34]{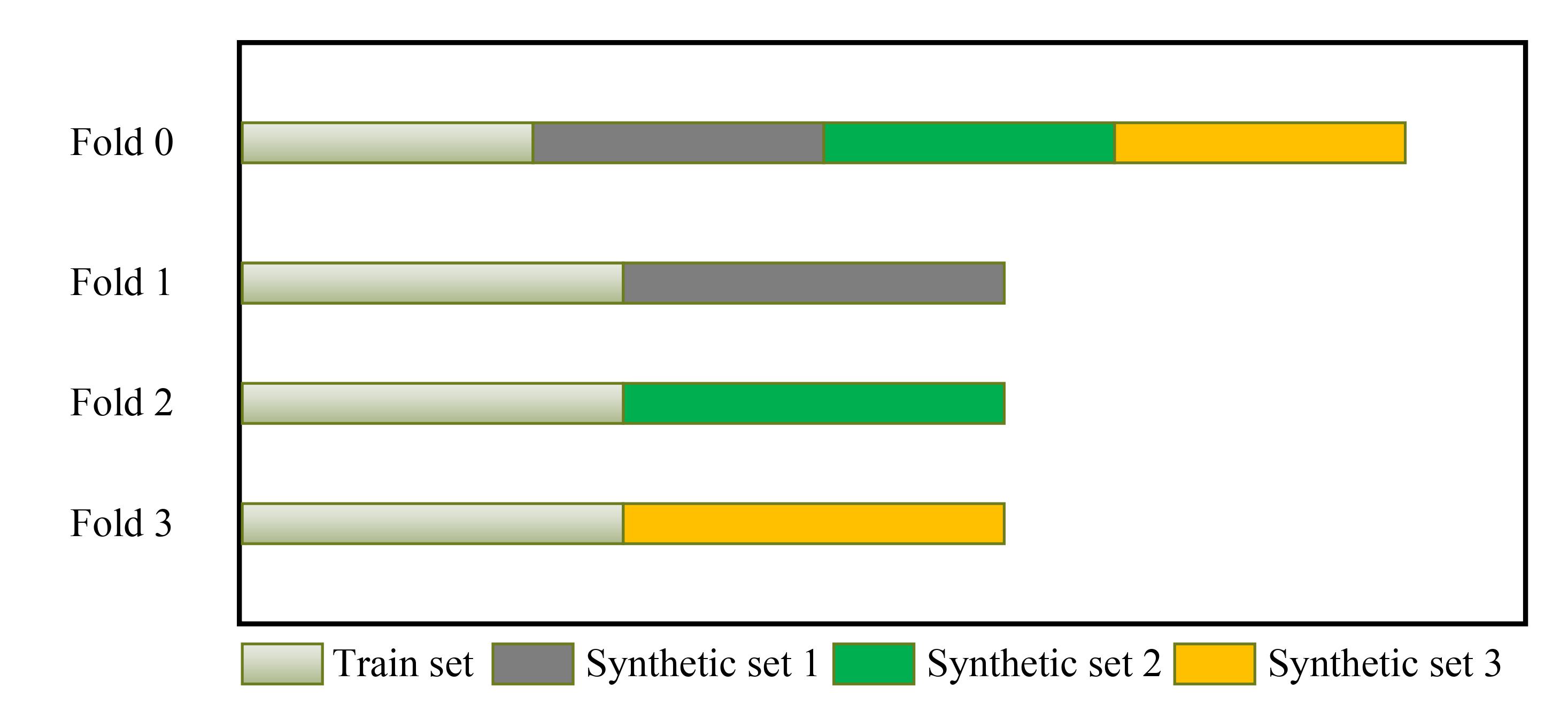}}
\vspace{0.2cm}
\caption{The proposed validation strategy for ensemble learning utilizing folds represented by K1, K2, and K3.}
	\label{FIG:5}
\end{figure}
\begin{table*}[t]
\centering
\caption{Performance evaluation using MSU-MFSD (M), Idiap (I), CASIA (C), and OULU-NPU (O) databases. The results are obtained directly from the cited papers.} \label{tab:cap3}
\setlength{\tabcolsep}{7.5pt}
\begin{tabular}{c | r | r | r | r | r | r | r | r }
\hline
& \multicolumn{2}{c |}{O\&C\&I to M} & \multicolumn{2}{c |}{O\&M\&I to C} & \multicolumn{2}{c |}{O\&C\&M to I} & \multicolumn{2}{c }{I\&C\&M to O} \\
\hline
Method & HTER & AUC & HTER & AUC & HTER & AUC & HTER & AUC \\
\hline
MADDG \cite{shao2019multi} & 17.69 & 88.06 & 24.50 & 84.51 & 22.19 & 84.99 & 27.89 & 80.02 \\
DAFL \cite{saha2020domain} & 14.58 & 92.58 & 17.41 & 90.12 & 15.13 & 95.76 & 14.72 & 93.08 \\
SSDG-R \cite{jia2020single} & 7.38 & 97.17 & 10.44 & 95.94 & 11.71 & 96.59 & 15.61 & 91.54 \\
DR-MD \cite{wang2020cross} & 17.02 & 90.10 & 19.68 & 87.43 & 20.87 & 86.72 & 25.02 & 81.47 \\
RFMetaFAS \cite{shao2020regularized} & 13.89 & 93.98 & 20.27 & 88.16 & 17.30 & 90.48 & 16.45 & 91.16 \\
FAS-DR-BC(MT) \cite{qin2021meta} & 11.67 & 93.09 & 18.44 & 89.67 & 11.93 & 94.95 & 16.23 & 91.18 \\
ADL \cite{liu2022adversarial} & \underline{5.00} & 97.58 & 10.00 & 96.85 & 12.07 & 94.68 & 13.45 & 94.43 \\
HFN + MP \cite{cai2022learning} & 5.24 & 97.28 & 9.11 & 96.09 & 15.35 & 90.67 & 12.40 & 94.26 \\
Cross-ADD \cite{huang2022generalized} & 11.64 & 95.27 & 17.51 & 89.98 & 15.08 & 91.92 & 14.27 & 93.04 \\
GDA \cite{zhou2022generative} & 9.20 & 98.00 & 12.20 & 93.00 & 10.00 & 96.00 & 14.40 & 92.60 \\
Regression Network \cite{kwak2023liveness} & 5.41 & \underline{98.85} & 10.05 & 94.27 & \underline{8.62}  & \textbf{97.60} & 11.42 & 95.52 \\
SSAN-R \cite{wang2022domain} & 6.67 & 98.75 & 10.00 & \underline{96.67} & 8.88 & \underline{96.79} & 13.72 & 93.63 \\
FG + HV \cite{liu2022feature} & 9.17 & 96.92 & 12.47 & 93.47 & 16.29 & 90.11 & 13.58 & 93.55 \\
CRFAS \cite{zheng2023learning} & 7.14 & 97.44 & 9.88 & 96.56 & \textbf{8.57} & 96.07 & 16.38 & 90.87 \\
\hline
CNN-LSTM w/Synt 1   &  8.65    & 98.02    & \underline{8.94}   & 97.05  &  38.37    & 91.53    & 10.60 & \textbf{99.22}\\ 
CNN-BiLSTM w/Synt 2 &  30.30   & 98.79    & 19.88   & 83.89  &  36.45    & 91.50     &  33.60 & 81.43 \\ 
CNN-GRU w/Synt 3    &  17.18   & 98.25    & 10.33   & 98.44  &   38.75    & 82.64      &  \underline{7.31} & 92.99 \\ 
Meta-Model (Stacked) &  \textbf{4.81}   & \textbf{99.19} & \textbf{8.92}    & \textbf{99.47}   &  38.11    & 94.45     & \textbf{6.70} & \underline{99.80} \\  
\hline
\end{tabular}
\end{table*}
\section{Experimental analysis of using open datasets}
To assess the effectiveness of the proposed use of synthesized face images, four publicly available databases are used: OULU-NPU database \cite{boulkenafet2017oulu} (denoted as O), CASIA Face Anti-Spoofing database (denoted as C) \cite{zhang2012face}, Idiap Replay-Attack database \cite{chingovska2012effectiveness} (denoted as I), and MSU Mobile Face Spoofing database \cite{wen2015face} (denoted as M). The performance is evaluated in-terms of Half Total Error Rate (HTER) (half of the sum of the false acceptance and false rejection rates) and Area Under Curve (AUC) on the target testing dataset.\\
\indent The OULU-NPU database contains 4950 videos, encompassing both real and attack videos. Two primary presentation attack types were considered in this database: print and video-replay. The videos recorded from the 55 subjects. CASIA-MFSD consists of a total of 50 subjects, with each subject having 12 videos captured under varying resolutions and lighting conditions. This dataset is developed to include three distinct types of spoof attacks: replay, warp print, and cut print attacks. In particular, CASIA-MFSD comprises 600 video recordings. Out of these, 240 videos from 20 subjects are allocated for training purposes, while the remaining 360 videos from 30 subjects are designated for testing. \\
\indent The Idiap Replay-Attack database comprises 1300 video clips that consist of photo and video attacks on 50 clients, all conducted under varying lighting conditions. Data for the attacks was gathered in two distinct lighting conditions: controlled, with office lights on, blinds down, and a uniform background; and adverse, with raised blinds, a complex background, and no office lighting. The MSU-MFSD dataset was constructed through the participants of 35 individuals, resulting in a combined total of 280 video entries. The recordings were captured using two distinct camera types, each with varying resolutions $(720\times480$ and $640\times480)$. Regarding generating live recordings, every participant contributed two video clips, one recorded with a laptop camera and the other with an Android device. In contrast, for video attack instances, two different camera models, specifically iPhone and Canon cameras, were utilized to capture high-definition videos for each subject. Thus, the presence of diverse lighting conditions and a wide array of attack types introduces significant complexity and difficulty when dealing with these datasets.

\subsection{Implementation details}
All the images are adjusted to a size of $224 \times 224$ to align with the input specifications of the pretrained DenseNet-201 architecture \cite{huang2017densely}. Fine-tuning of the CNN model is carried out employing the Stochastic Gradient Descent (SGD) optimizer, with a mini-batch size of $32$ and a validation check performed every $30$ iterations. The learning rate is set at $0.0001$, and fixed-size epochs are not used. Instead, we implement an early stopping mechanism \cite{prechelt1998early} to automatically stop the training process to prevent overfitting. During the ensemble learning phase, the CNN model undergoes fine-tuning with spatiotemporal-encoded video clips and three distinct synthetic sets individually. These subsets introduce random expansion to the training images through alpha values of $0.5$, $1.0$, and $1.5$ as defined in Eq. 4. Subsequently, the features extracted from each of these fine-tuned models are separately used as inputs to train a LSTM, a BiLSTM and a GRU. 

\begin{table*}[t]
\centering
\caption{Base model architectures and their parameters.}
\begin{tabular}{c c c c c} 
 \hline
  &  Base model 1 & Base model 2  & Base model 3  & Meta-Model  \\ 
  \hline
Layer type & LSTM  & BiLSTM  & GRU & GRU \\ 
\hline
No. of layers & 1  & 1 & 1 & 1 \\ 
 \hline
No. of units & 500  & 20  & 20 & 100 \\ 
 \hline
Optimizer & ADAM & ADAM & ADAM & ADAM  \\ 
 \hline
Learning rate & 0.0001 & 0.0001 & 0.0001  & 1\\ 
 \hline
Cost function & cross entropy & cross entropy & cross entropy & cross entropy  \\ 
 \hline
\end{tabular}
\end{table*}

\begin{table*}[t]
\centering
\caption{The results of cross-dataset testing on limited source domains. The results are obtained directly from the cited papers.}
\label{tab:cap4}
\begin{tabular}{c|r|r|r|r|r|r|r|r|r|r}
\hline
 & \multicolumn{2}{c|}{O\&I to M} & \multicolumn{2}{c|}{M\&I to C}  & \multicolumn{2}{c|}{O\&I to C} & \multicolumn{2}{c|}{O\&M to I} & \multicolumn{2}{c}{C\&M to O}\\ \hline
  Method  & HTER       & AUC    & HTER   & AUC & HTER         & AUC        & HTER    & AUC    & HTER     & AUC   \\ \hline
Supervised \cite{wang2021consistency} &  12.1 &  94.2  & 30.4 & 77.0 & 18.0 &  90.1 & 16.8 & 93.8 & 17.9 & 89.5 \\ 
Mean-Teacher \cite{tarvainen2017mean} & 19.6  & 86.5  &  31.1 &   76.6 &   23.7 &   84.9  & 18.4 &  86.0  & 23.5  &  84.9 \\ 
USDAN  \cite{jia2021unified} & 15.8  & 88.1  & 35.6 & 69.0 & 33.3 & 72.7 & 19.8 & 87.9   & 20.2  &  88.3\\ 
EPCR-labeled \cite{wang2021consistency} & 12.5  & \underline{95.3}& \underline{18.9} & 89.7 & 18.9 & 89.7 & \underline{14.0} & 92.4  & 17.9  & 90.9 \\ 
EPCR-unlabeled  \cite{wang2021consistency}  & \underline{10.4}  & 94.5 & 25.4 & 83.8  & 16.7 & \underline{91.4}& \textbf {12.4} & \underline{94.3} & 17.8  &   \underline{91.3}  \\
           \hline 
CNN-LSTM w/Synt 1  & 19.6  & 86.5  &  31.1 &  76.6 &  23.7 &  84.9 & 22.4 & 84.0 & 23.5 &  84.9 \\ 
CNN-BiLSTM  w/Synt 2  & 15.8  & 88.1  & 35.6 & 69.0 & 33.3 & 72.7 & 19.8  & 87.9  & 20.2  &  88.3\\ 
CNN-GRU w/Synt 3 & 14.1 & 94.8 & 19.3 & \underline{90.3} & \underline{15.4}& 88.2 & 20.1 & 90.2  &  \underline{14.9} & 91.2 \\ 
Meta-Model &  \textbf {8.1} & \textbf {99.1}  & \textbf {18.3} & \textbf {92.7} &  \textbf{11.2} & \textbf{97.5} & 14.5 & \textbf {94.7}  & \textbf{14.1}  & \textbf{93.9}  \\ 
\hline
\end{tabular}
\end{table*}

\begin{table*}[t]
\centering
\caption{Ablation study using cross-database evaluation.}
\label{tab:cap5}
\begin{tabular}{c|r|r|r|r|r|r|r|r}
\hline
 & \multicolumn{2}{c|}{O\&C\&I to M} & \multicolumn{2}{c|}{O\&M\&I to C}  & \multicolumn{2}{c|}{O\&C\&M to I} & \multicolumn{2}{c}{I\&C\&M to O} \\ \hline
  Method  & HTER & AUC & HTER & AUC & HTER & AUC  & HTER  & AUC    \\ \hline
CNN w/o synthetic data &  23.70   & 83.13    & 24.01   & 90.34  &  39.44  & 80.47   &  37.55 & 84.30 \\ \hline
CNN-LSTM w/Synt 1   &  8.65    & 98.02    & 8.94    & 97.05  &  38.37    & 91.53    & 10.60 & 99.22\\ 
CNN-LSTM w/Synt 2      &  17.96   & 89.86    & 12.10   & 95.82  &  38.74  & 79.75   &  12.55 & 99.71 \\ 
CNN-LSTM w/Synt 3      &  21.25   & 92.05    & 11.79   & 95.41  &  37.99  & 71.52   &  8.44 & 99.76 \\ \hline
Base model 1            &  5.63   & 98.00    & 9.65    & 98.28  &  36.67   & 82.72   &  5.35 & 99.79 \\ 
\hline\hline
CNN-BiLSTM w/Synt 1 &  8.28   & 99.93    & 18.03   & 89.08  &  36.18 & 99.07   &  25.85 & 99.85 \\ 
CNN-BiLSTM w/Synt 2 &  30.30   & 98.79    & 19.88   & 83.89  &  36.45    & 91.50     &  33.60 & 81.43 \\ 
CNN-BiLSTM w/Synt 3 &  21.03   & 97.60    & 18.89   & 90.38  &  34.63 & 97.52   &  32.93 & 90.54\\ \hline
Base model 2         &  3.95   &  99.97    & 15.46   & 82.09  &  33.12 & 61.60   &  25.41 & 97.00 \\ 
\hline\hline
CNN-GRU w/Synt 1 &  9.34   & 99.71    & 20.21   & 89.58  &  39.18 & 86.22   &  26.88 & 98.13 \\ 
CNN-GRU w/Synt 2 &  21.07   & 98.37    & 15.63   & 95.93  &  39.30 & 80.73   &  6.15 & 98.38 \\ 
CNN-GRU w/Synt 3    &  17.18   & 98.25    & 10.33   & 98.44  &   38.75    & 82.64      &  7.31 & 92.99 \\ \hline
Base model 3       &  14.44   & 99.80    & 14.72   & 98.06  &  39.67 & 85.52   &  8.16 & 92.28 \\ 
\hline\hline
\end{tabular}
\end{table*}

\subsection{Comparison against the state-of-the-art methods}
To compare the performance with recently introduced domain generalization methods, we conduct cross-dataset testing, where the model is trained on three source databases and evaluated on a completely unseen database using the leave-one-out (LOO) strategy. In particular, the testing sets of source databases are used as a validation set for computing the Equal Error Rate (EER). Thus, the Half Total Error Rate (HTER) is calculated directly on the target (unseen) dataset. As shown in Table I, the proposed ensemble learning provides the best results on three protocols: O\&C\&I to M, O\&M\&I to C, I\&C\&M to O. The meta-model demonstrates its effectiveness when the predictions from three different classifiers, such as  LSTM, BiLSTM, and GRU, are combined using our GRU-based meta-model. Thus, the meta-model plays a crucial role in enhancing the generalization ability of the face anti-spoofing task. The learning parameters used for training different base models are described in Table II.\\
\indent It is worth mentioning here that we accumulate the output of the three base models' validation sets as the new validation set for training the meta-model. This way, the meta-model makes the final test prediction on the test set. We argue that the proposed data augmentation technique facilitates the base models in learning from the diversity within the data. Consequently, temporal inconsistencies or blurriness prove to be valuable in distinguishing genuine images from those associated with attacks. In contrast to previous approaches that have employed adversarial learning \cite{liu2022adversarial}, generative domain adaptation \cite{zhou2022generative}, and meta-learning \cite{cai2022learning}, it becomes apparent that ensemble learning, particularly when employing a stacking-based approach, achieves notably competitive performance.\\
\indent We also provide a more comprehensive evaluation of the classifier's performance using the Area Under the ROC Curve (AUC) shown in Table I. Although the meta-model provides low performance on one database (i.e., Replay-Attack), one can see that the meta-model achieves more than 90\% AUC on all the datasets. Since EER is calculated based on the testing set of source databases, HTER focuses on finding the operating point (threshold) where the False Acceptance Rate (FAR) and False Rejection Rate (FRR) are equal. If the AUC is high and the HTER is low, it indicates that the classifier achieves a good balance between FAR and FRR. Since AUC is not threshold-dependent, it is more useful for comparing different classifiers or evaluating the model's generalization capability.

\subsection{Experiment on Limited Source Domains}
Our study also investigates the scenario of a limited source domain by training the model on only two source domains, as opposed to the three domains mentioned in Table III. It can be observed that the model continues to exhibit superior performance across the target domains. In particular, the model achieves the lowest Half Total Error Rate (HTER) in four protocols and the highest Area Under the Curve (AUC) score on all target domains. This outcome emphasizes that even with limited source data, the stacking ensemble approach maintains its robust generalization capability. Overall, this finding is significant as it demonstrates the effectiveness of the stacking ensemble approach in adapting to scenarios where a limited amount of source data is available.
\begin{table*}[t]
\centering
\caption{Average execution time in seconds.} \label{tab:cap7}
\begin{tabular}{c | r | r | r | r}
    Dataset & Optical flow \cite{horn1981determining} & ASGS method \cite{muhammad2022adaptive} & TSS method \cite{muhammad2022self} & Ours
  \\
  \hline
     CASIA-FASD  & 1560 & 1487 & 1140 &  1023 \\
      REPLAY-ATTACK & 1082 & 1003 & 780 & 641\\  
  \hline
\end{tabular}%
\end{table*} 
\begin{figure*}
	\centering
	\includegraphics[scale=.35]{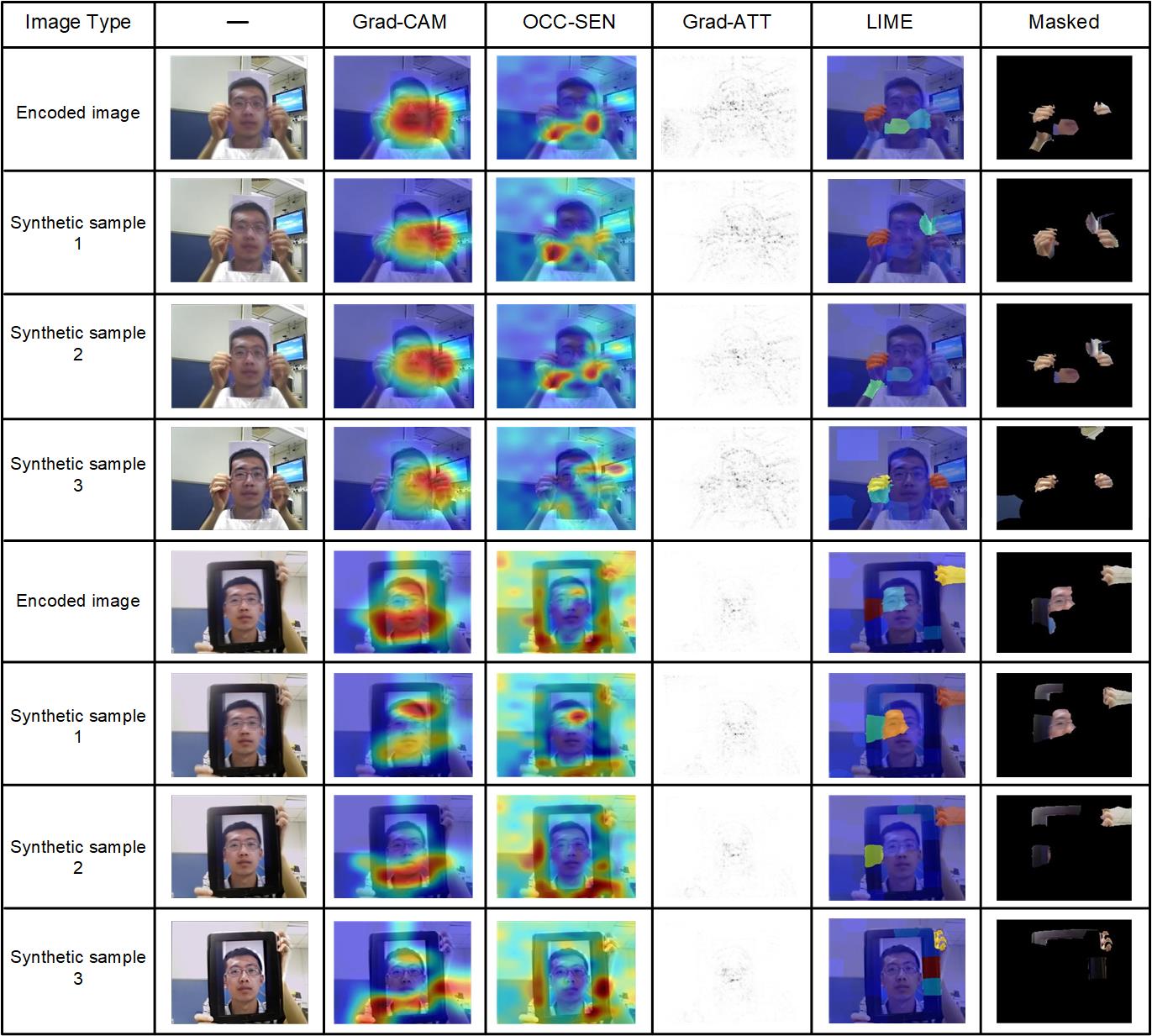}
	\vspace{0.3cm}
	\caption{Displaying feature maps through visualization. The types of images are labelled in the first column. The second column shows the original encoded and synthetic images. The third column illustrates the feature maps from Grad-CAM \cite{selvaraju2017grad} while the fourth column shows the feature maps from occlusion sensitivity maps \cite{zeiler2014visualizing}. Similarly, the fifth and sixth column visualize the features maps from Gradient Attribution map using Guided Backpropagation \cite{springenberg2014striving}, and locally interpretable model-agnostic explanations \cite{ribeiro2016should}, respectively. The last column shows the masked images obtained from LIME predictions.}
	\label{FIG:6}
\end{figure*}
\subsection{Ablation study}
In order to assess the effectiveness of our ensemble learning method and analyze the individual contributions of each base model, we conducted separate experiments in multi-source domains. The results of these experiments are detailed in Table IV, focusing primarily on multi-source domain scenarios. The baseline results represent the performance of the fine-tuned DenseNet-201 architecture without incorporating synthetic data. These results are based on spatiotemporal-encoded images obtained through the video distillation process. Next, we demonstrate how incorporating a GRU-based meta-model enhances the performance of individual predictions generated by LSTM, BiLSTM, and GRU networks. Starting with the first base model, the LSTM model is trained using the original CNN spatiotemporal-encoded features, along with features extracted from three synthetic subsets. After obtaining predictions from the LSTM model for each subset, we horizontally concatenate them and utilize a meta-model based on the GRU network. This meta-model is designed to make final detection by leveraging the combined knowledge gathered from each subset of synthetic samples generated by the LSTM models.\\
\indent In a similar manner, we developed the other two base models using the BiLSTM and GRU networks. By doing so, we employ a GRU-based meta-model to combine the predictions from each subset of synthetic images. Based on our experimental analysis, we consistently observe that including synthetic images leads to increased diversity in performance across all datasets. Moreover, incorporating a GRU-based meta-model to combine the predictions of each base model further strengthens the model's robustness, especially on datasets that were previously unseen. Hence, these experimental results illustrate the importance of the proposed data augmentation strategy and emphasize how the introduction of a meta-model further enhances the method's generalization capabilities.\\
\indent {\em Comparisons of execution times:}
We analyze the execution times of the proposed video distillation technique with the previous global motion estimation methods \cite{muhammad2022self, muhammad2022adaptive} and optical flow\cite{horn1981determining}. Table V reports the numerical results in the total number of seconds used to generate the training samples on two datasets. All these comparison results were reported by using a MATLAB environment based on a workstation with 3.5 GHz Intel Core i7-5930k and 64 GB RAM. One can see that the proposed global motion estimation technique is computationally less expensive than the previous motion estimated methods reported recently in the literature. This is due to the fact that the FAST \cite{rosten2005fusing} feature detector is designed with a focus on computational efficiency and speed, while FREAK \cite{alahi2012freak} is intended to work in combination with fast feature detectors like FAST, providing a matching mechanism that is both fast and robust.
\begin{figure*}
    \centering
    \begin{subfigure}[b]{0.3\textwidth}
        \centering
        \includegraphics[width=\textwidth]{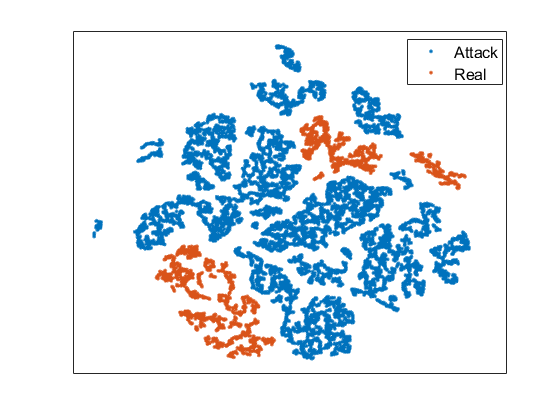}
        \caption{}
    \end{subfigure}
    \hfill
    \begin{subfigure}[b]{0.3\textwidth}
        \centering
        \includegraphics[width=\textwidth]{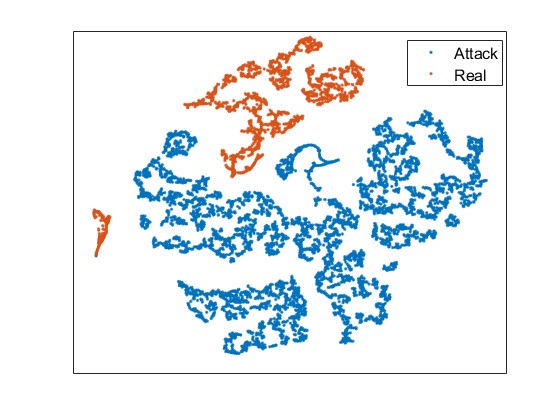}
        \caption{}
    \end{subfigure}
    \hfill
    \begin{subfigure}[b]{0.3\textwidth}
        \centering
        \includegraphics[width=\textwidth]{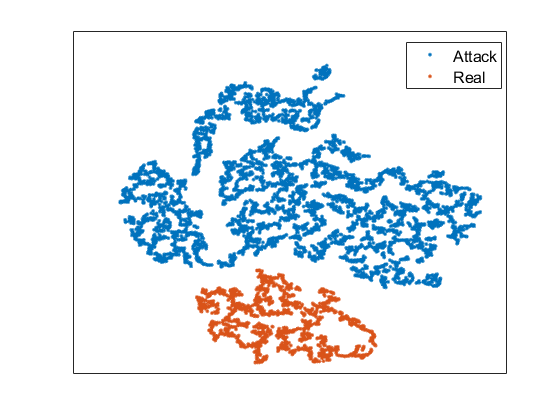}
        \caption{}
    \end{subfigure}
    \caption{The T-SNE visualization feature distributions on cross-testing scenarios. (a) shows the feature distribution of the original encoded video clips, (b) reflects the feature distribution of encoded video clips with a subset of synthetic samples, (c) shows the feature distribution of the meta-learner.}
    \label{FIG:7}
\end{figure*}
\begin{figure*}
   \centering
       \begin{subfigure}[b]{0.2\textwidth}
        \centering
          \includegraphics[width=\textwidth]{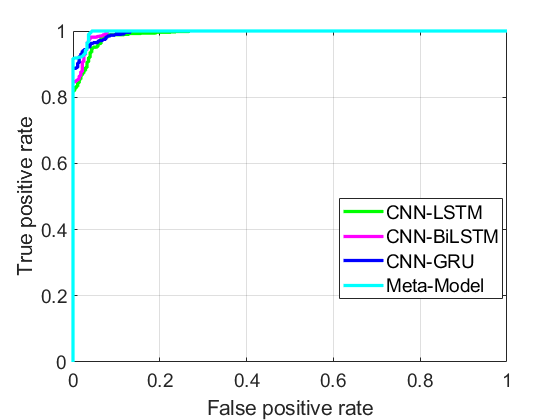}
         \caption{}
     \end{subfigure}
     \hfill
     \begin{subfigure}[b]{0.2\textwidth}
        \centering
          \includegraphics[width=\textwidth]{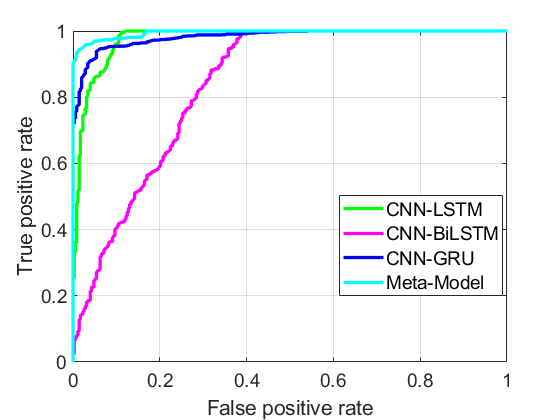}
         \caption{}
     \end{subfigure}
     \hfill
     \begin{subfigure}[b]{0.2\textwidth}
        \centering
         \includegraphics[width=\textwidth]{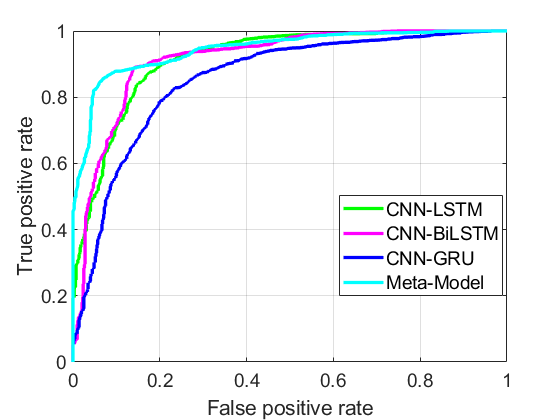}
         \caption{}
     \end{subfigure}
      \hfill
     \begin{subfigure}[b]{0.2\textwidth}
        \centering
          \includegraphics[width=\textwidth]{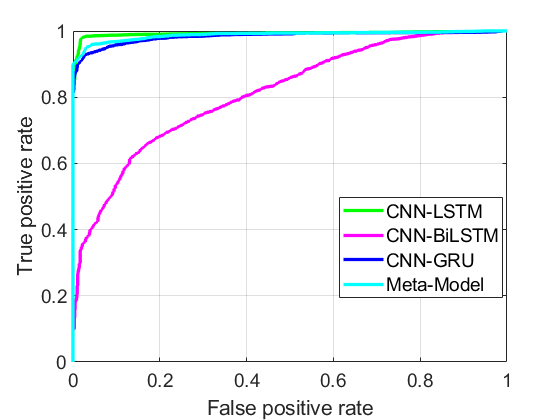}
         \caption{}
     \end{subfigure}
          \hfill
        \caption{The Receiver Operating Characteristics (ROC) curves. (a) O\&C\&I to M, (b) O\&M\&I to C, (c) O\&C\&M to I, and (d) I\&C\&M to O are developed by plotting the true positive rate (TPR) against the false positive rate (FPR).\label{fig8}}
\end{figure*}

\subsection{Interpretation of the deep neural network}
Interpretation is essential to observe the learning patterns in data that are important, but there is no clear consensus on how interpretability should be best defined in the context of machine learning \cite{molnar2020interpretable}. Although explanation methods intend to make neural networks more trustworthy and interpretable, the “black-box” nature of deep neural networks can make it challenging to determine precisely why a particular decision was made. For instance, synthetic samples provide additional variations of the data and lead to better interpretability compared to the same model trained without synthetic samples. This improvement is attributed to the fact that the motion cues, which are naturally available in the frame sequences, are “easy to learn” for the model and play an important role in model optimization. Consequently, the importance of interpretation is becoming increasingly popular and has led to useful and promising findings. \\
\indent In our work, we utilize Gradient-weighted Class Activation Mapping (denoted as Grad-CAM) \cite{selvaraju2017grad}, Occlusion Sensitivity Maps (denoted as OCC-SEN) \cite{zeiler2014visualizing}, Gradient Attribution Map using Guided Backpropagation (denoted as Grad-ATT) \cite{springenberg2014striving}, and Locally Interpretable Model-Agnostic Explanations (denoted as LIME) \cite{ribeiro2016should} to understand which patterns in the data are deemed important and contribute to the final decision. These methods enable us to trust the behavior of the developed deep learning model and/or further tune the model by observing its interpretations. Specifically, we extract visualization maps from the pretrained DenseNet-201 \cite{huang2017densely} convolutional neural network for all of the above methods in our experiments. In Fig. 6, we visualize diverse sets of synthetic images from the CASIA dataset. The first four rows show print attack images, while the next four rows show replay attack images. Each visualization method captures the class discriminative region, thanks to the proposed synthetic data generation scheme that allows the network to use more subtle cues for its correct classification. In particular, the first row shows that the neurons in the deep convolutional layers focus on the paper's texture and hand movement cues. However, Grad-ATT \cite{springenberg2014striving} interpretation shows that the model also takes the background as context to make the prediction. Surprisingly, this issue is eliminated by the proposed synthetic data generation scheme, where the second, third, and fourth rows show that the model only considers motion cues, surface edges, and barely observes the background context. \\
\indent In the case of a replay attack, the remaining rows show that the tablet screen and hand movement provide discriminative information for the model's prediction. While we cannot present this for every image of the dataset, we observed that the mouth region, eye blinking, and head rotation contribute positively to distinguishing live and spoofed images. Thus, interpretations from the above methods demonstrate that the proposed learning model is focusing on the correct features of the input data, and the model's decision can be viewed in a human-understandable way. Moreover, the proposed synthetic data generation method provides informative RGB images and helps the model to make the features of spoofed faces more dispersed, allowing a better class discrimination to generalize well to the target domain.

\subsection{Visualization and Analysis}
In order to visually illustrate the individual contributions of each model, we employ t-SNE (t-Distributed Stochastic Neighbor Embedding) to analyze the distribution of different features, as depicted in Fig. 7. Initially, the model is trained on the 0+C+I source domains without incorporating synthetic samples, resulting in a trivial distribution shown in Fig. 7(a). In this representation, the boundary between live and spoofed samples is indistinct, and areas of overlap can lead to potential misclassifications, thereby degrading overall performance.\\
\indent However, when synthetic samples are included in the model, as depicted in Fig. 7(b), the feature distribution demonstrates improvement, offering a comparatively clearer separation compared to the model that does not include synthetic samples. This enhancement is attributed to the synthetic samples aiding the model in recognizing spatiotemporal artifacts. Nonetheless, with the introduction of the meta-model, we observe a well-structured and compact distribution with a clearly defined boundary in Fig. 7(c). As a result, our proposed ensemble learning approach exhibits strong generalizability when applied to unseen target data.\\
\indent In Fig. 8, we employ ROC curves to visually represent the model's ability to differentiate between real and attack classes. As demonstrated in Fig. 8, the meta-model consistently achieves an AUC (Area Under the Curve) of over 90\% across all datasets, showcasing an impressive level of performance on previously unseen testing sets. The ROC curve is constructed with True Positive Rate (TPR) plotted against False Positive Rate (FPR), where FPR is on the x-axis and TPR is on the y-axis. Specifically, when the meta-model (ensemble) shifts the curves closer to the top-left corner, it indicates superior performance in distinguishing between the classes.

\section{Conclusions}
In this paper, we addressed the domain generalization issue in face presentation attack detection (PAD) by proposing a novel approach that combines data augmentation and deep ensemble learning. By observing multiple blurred sequences, the base models were able to learn sequential patterns of motion blur and infer how objects move over time. This data augmentation technique is found to be helpful in improving the robustness of the models and their ability to handle variations in motion blur present in real-world scenarios. This technique not only enhanced the training data but also addressed the issue of black framing that might arise during feature-based global motion estimation. Based on the experimental results, the performance of LSTM, BiLSTM, and GRU still faces limitations in certain scenarios. To improve the overall generalization, we introduced a meta-model that leverages the strengths of different base models. This ensemble approach allowed the model to benefit from the diverse representations learned by individual base models. Based on the experimental results on four benchmark datasets, the meta-model achieves state-of-the-art performance on three datasets.\\
\indent Finally, the interpretation of the model shows that motion cues (e.g., temporal information or motion patterns) are helpful in improving the model's generalization ability. We conclude that the effectiveness of a meta-model depends on the diversity and quality of the base models used. If the base models suffer from similar limitations or biases, the meta-model may not provide significant improvements. Our future work will focus on the development of robust motion estimation methods in end-to-end learning to improve the generalization of face PAD.

\section{Declaration of Competing Interest}
The authors have no conflict of interest that could have appeared to influence the work reported in this paper.

\section{Acknowledgments}
This work is financially supported by ‘Understanding speech and scene with ears and eyes (USSEE)” (project number 345791). The first author also acknowledges the support of the Ella and Georg Ehrnrooth foundation. 

\ifCLASSOPTIONcaptionsoff
  \newpage
\fi

\bibliographystyle{ieeetr}
\bibliography{cas-refs}

\end{document}